\documentclass[sigconf]{acmart} 

\AtBeginDocument{%
  \providecommand\BibTeX{{%
    \normalfont B\kern-0.5em{\scshape i\kern-0.25em b}\kern-0.8em\TeX}}}

\setcopyright{acmcopyright}
\copyrightyear{2021}
\acmYear{2021}
\acmDOI{10.1145/1122445.1122456}

\acmConference[KDD'21]{KDD}{August 14--18, 2021}{Singapore}
\acmBooktitle{KDD'21: Knowledge Discovery and Data Mining}
\acmPrice{15.00}
\acmISBN{978-1-4503-XXXX-X/18/06}

\usepackage{tikz}
\usetikzlibrary{arrows.meta,positioning}
\usepackage{booktabs} 
\usepackage{amsthm}
\usepackage[utf8]{inputenc}
\usepackage{graphicx,fancyhdr,amsmath,amsthm,url,hyperref, verbatim}
\usepackage{subcaption}
\usepackage{algorithmic}
\usepackage{algorithm}
\usepackage{wrapfig}
\usepackage{tikz}
 \usetikzlibrary{calc} 

\usepackage{bbm, enumerate, setspace,pdfpages,tabularx,lipsum,environ }
\usetikzlibrary{arrows.meta,positioning}
\usepackage{tablefootnote}
\allowdisplaybreaks
\usepackage[section]{placeins}
\setcopyright{acmcopyright}

\usepackage[disable]{todonotes}

\newcommand{\nicocom}[1]{\todo[inline, color=purple]{Nico: #1}}

\newcommand{\paricom}[1]{\todo[inline, color=yellow]{Pari: #1}}

\begin{document}

\title{QUEST: Queue Simulation for Content Moderation at Scale}

\author{Rahul Makhijani}
\affiliation{%
  \institution{Core Data Science, Facebook}
}
\author{Parikshit Shah}
\affiliation{%
  \institution{Core Data Science, Facebook}
}
\author{Vashist Avadhanula}
\affiliation{%
  \institution{Core Data Science, Facebook}
}
\author{Caner Gocmen}
\affiliation{%
  \institution{Core Data Science, Facebook}
}
\author{Nicol\'{a}s E. Stier-Moses}
\affiliation{%
  \institution{Core Data Science, Facebook}
}
\author{Juli\'{a}n Mestre}
\affiliation{%
  \institution{Core Data Science, Facebook}
}

\renewcommand{\shortauthors}{Makhijani et al.}
\renewcommand{\shorttitle}{QUEST: A Queue Simulation Tool}

\begin{abstract}
  Moderating content in social media platforms is a formidable challenge due to the unprecedented scale of such systems, which typically handle billions of posts per day. Some of the largest platforms such as Facebook blend machine learning with manual review of platform content by thousands of reviewers. Operating a large-scale human review system poses interesting and challenging methodological questions that can be addressed with operations research techniques. We investigate the problem of optimally operating such a review system at scale using ideas from queueing theory and simulation.
\end{abstract}



\maketitle

\section{Introduction}
\label{sec:introduction}

Content moderation is a central problem faced by many online platforms. This is especially important in social media platforms such as Facebook and Instagram. Moderation typically involves analyzing content shared on the platform to detect and remove those that violate the platform's policies (e.g., posts containing harmful content). One particular challenge in content moderation is the sheer scale and complexity that these systems must cope with. For example, in Q3 2020 alone, the prevalence of hate speech (fraction of content views that violate community standards on hate speech) was $0.1$-$0.11\%$ and an action was taken on roughly 57 millions pieces of content~\cite{FacebookCSERnov20}. Although advances in artificial intelligence have helped automating some aspects of content moderation, Facebook still relies on human reviewers for moderating ``ambiguous content'' that is harder for computers to classify reliably. For instance, discerning if someone is the target of bullying on Facebook can be extremely nuanced and contextual, and therefore requires human judgement in accurately adjudicating if a flagged post violates its Community Standards~\cite{FacebookCS,FacebookCM}. While such hybrid systems can help Facebook handle content moderation at scale, it also brings up a number of operational challenges. 

Facebook builds and maintains its content moderation system based on both automation and human reviews. On the automation side, this involves building and deploying machine learning (ML) models that monitor content. The human content moderation faces different challenges such as forecasting staffing requirements, and deciding how to allocate the workforce's time between content labeling to train ML models and policy enforcement. In this paper, we focus on the challenges of operating such a large scale human content moderation system and the tools Facebook has developed to tackle these challenges.



In order to describe the setting in more detail, we first introduce some terminology:



\begin{itemize}
\item A {\bf job} represents a reviewing task associated with a specific piece of content. Jobs are generated every time a user reports a piece of content, or when an ML classifier flags content as potentially violating policy, or might be automatically derived from an existing job. For example, if a user reports a page for some specific type of violation, we might fan out multiple ``disaggregated jobs'' for individual posts on the original page.

Jobs have intrinsic attributes such as content type (text, image, video, etc), language, market (geographical region where the content originates), and suspected violation type (hate speech, nudity, etc). It is possible for multiple jobs to point to the same content; e.g., a single post flagged by two different users at two different times would lead to the creation of two different jobs. 

\item A {\bf reviewer} is a person who works on a job and makes the decision regarding whether to classify it as benign, pass on the job to a different reviewer, or take a enforcement action (such as removing it from the platform). 

\item A {\bf queue} is a container that holds a pool of jobs with the similar attributes (e.g. language, potential violation type, etc.). Jobs are assigned to queues based on their attributes; for example, a queue may comprise of jobs related to hate speech violations for content generated in a particular market.




\item {\bf Turn Around Time (TAT)}:
 is defined as the total elapsed time between when a job is enqueued and when it is finally decisioned. 

 \item {\bf Service Level Agreement (SLA)}:
 is a contract regarding TAT for jobs, i.e. a specification of an upper bound on how long the system may take to review jobs. 

 \end{itemize}


\begin{figure}
  \centering
  \begin{tikzpicture}[status/.style={font=\small,draw,fill=yellow!20,rectangle,rounded corners,inner sep=6pt,font=\footnotesize}, edge_label/.style={fill=white,font=\footnotesize},>=Stealth,scale=0.7] 
    \node[status] (create) at (0,0) {Create};
    \node[status] (route) at (1, 3) {Route to Queue};
    \node[status] (assign) at (5, 3) {Assign Reviewer};
    \node[status] (decide) at (6, 0) {Decide};
    \node[status] (close) at (10, 0) {Close};
    \draw[->] (create) -- node[edge_label] {Human Review} (route);
    \draw[->] (create) -- node[edge_label] {ML Review} (decide);
    \draw[->] (route) -- (assign);
    \draw[->] (assign) -- (decide);
    \draw[->] (decide) -- (close);
    \draw[->] (decide) -- (route);
  \end{tikzpicture}
  \vspace{1em}
\caption{Workflow of a job}
  \label{fig:workflow}
\end{figure}
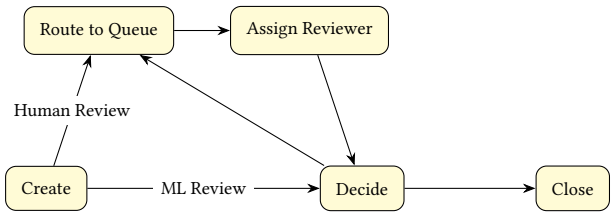

Fig.~\ref{fig:workflow} gives a high level overview of a job's life cycle. The system resembles a queueing network \cite{queueingnetworks}, which are thoroughly studied in the context of call centers \cite{journals/anor/KooleM02}. Jobs are analogous to incoming calls and reviewers to call center agents. However, two key differences between the call center and the content moderation settings are \emph{scale} and \emph{routing complexity}. Regarding scale, we note that typical call centers \cite{call-centers} have hundreds of employees who take many minutes to serve each job, while Facebook contracts thousands of reviewers who's job action time can vary from a few seconds to many minutes. Regarding complexity, note that call centers all information about a job is rather simple and known at the time of arrival (e.g., type of enquiry, client priority, etc), whereas routing in content moderation needs to take into account more complex factors that are unique to each piece of content like ML classifier confidence scores and dynamic attributes like spread and virality that change in real time.

Despite the extensive literature on queueing theory, our real-world application exhibits a number of nuances that violate the standard assumptions needed by these theoretical frameworks. As a concrete example, we consider the work by Mandelbaum and Stolyar~\cite{cmu}, where they analyze the optimality of a class of policies know as ``c-mu'' rules relying on the following assumptions:

\begin{enumerate}
    \item Statistical assumptions such as the system being stationary (i.e., arrival and service rates do not change over time) and well-provisioned (i.e., arrival rates are commensurate with processing rates). Neither of these hold in our case. 
    \item The ``c-mu'' literature typically assumes that cost functions are convex and increasing, and known when the job arrives. This is not the case is our setting since the cost depends on the number of content views (which need not be convex) and the content severity (which is not known until it is reviewed).
    \item Furthermore, the ``c-mu'' rule assumes differentiable cost functions. In practice, content views are point processes and thus one needs to work with finite differences (as opposed to point derivatives). The literature provides no guidance on how to size these finite differences, i.e., what the window-size should be.
\end{enumerate}

These challenges make it impossible to rely on closed-form solutions from queuing theory that would enable us to derive precise estimates for various system-level metrics (e.g., for the average job TAT). In practice, these limitation are overcome by building detailed simulation models and relying on their outputs to estimate relevant metrics \cite{callcenter}. Simulation models are commonly used to understand and optimize the operations of complex systems related to manufacturing, call centers, healthcare and bike sharing \cite{simulationbook, bikesimulation, gunal2010simulation}.

In this paper we describe QUEST, a new simulation tool used at Facebook that is able to tackle the operational challenges arising in human content moderation systems at scale. One significant challenge we address is evaluating the impact of potential decisions (e.g., hiring additional reviewers or training them in new skills) without actually having to make them. Most changes require long term planning and cannot be easily reverted if the decision turns out to be not as expected. Running A/B tests is also challenging due to network effects, since each reviewer is eligible to work on multiple queues, which would be subject to spillover effects.

The rest of this paper is organized as follows. In Sect.~\ref{sec:overview}, we cover the details of QUEST including its inputs, outputs, performance evaluation and some domain specific challenges. The next section, Sect.~\ref{sec:usecases}, discusses how simulator may be used to derive operational decisions at Facebook; specifically, we talk about leveraging the simulator to address capacity planning, automation, queue prioritization and supply-demand matching problems. Lastly, in Sect.~\ref{sec:conclusion} we give a summary of potential future extensions of our work. As a general caveat, it is worth noting that the plots in the paper are generated based on simulated data, and the y-axes numbers are obfuscated to prevent revealing company-sensitive information, and merely hold directional information.

\subsection{Related work}

As content moderation has risen in importance over the last decade across social media platforms, it has gotten significant attention in academia and popular media. \citet{jhaverThesis} gives an overview of these systems. As mentioned earlier, a common aspect across content-moderation and crowd-sourcing platforms is that they leverage both automation and human review \cite{youtubeMashable, youtubeVerge, jhaver2019human}. A recent overview of how these systems work together at Facebook can be found in \cite{FacebookCM, facebookVerge}. In addition to traditional binary or multi-class prediction models that the company relies on, Facebook has also invested in two specialized systems that are particularly useful for content moderation.
CLARA is a sophisticated statistical framework developed and deployed at Facebook that takes as input the decisions of reviewers and relying on ML algorithms, outputs the prevalence of violating content and the confusion matrix of reviewers \cite{clara}.
This is used to aggregate the decisions make by several reviewers efficiently thereby increasing the accuracy. 
Another important and related framework developed at Facebook is a machine learning system that can 
predict the popularity of social network content over arbitrary time horizons, given information about
the content's initial popularity growth and other content features \cite{vespa}. These forecasts could be used to prioritize content sent to human review if one wants those priorities to depend on anticipated distribution.

To the best of our knowledge, our paper is the first documented application of simulations models to content moderation systems.
Besides the specifics of the application domain, the main difference to other queuing system models is scale since these systems involve a large number of reviewers that need to provide content moderation for billions of content pieces per day. To give an idea of the scale, Facebook took action on 
22.5 million of pieces in the second quarter of 2020 \cite{FacebookCSERaug20}, and 
22.1 million pieces of hate speech pieces 
and
19.2 million pieces of violent and graphic content
in the third quarter of the same year, in addition to other areas and more pieces on Instagram
\cite{FacebookCSERnov20}.
The metrics of interest in this system are different from other simulations, as discussed later.

\section{QUEST Overview}
\label{sec:overview}

The QUEST framework is based on a discrete event simulator that replicates the operations of the queues, reviewers and jobs. 
We built the system to visualize the counterfactual evaluation of changes  that enables us to answer what-if questions. At a high level, the process involves these steps:

\begin{itemize}
\item \textbf{Gathering inputs}: The initial step is to gather historical data about jobs, queues, reviewers and their schedules.
\item \textbf{Simulation}: After the setup is done, we perform the job and event level simulation of the real-life workflows.
\item \textbf{Evaluation metrics}: Finally, we compute relevant metrics on both historical as well as simulated data, which we use to make decisions.
\end{itemize}

The next sections describe each of the previous bullets in detail.

\subsection{Simulation Inputs}

The simulation is complex and is based on several input streams, capturing supply, demand, and queue structure and possibly a counterfactual setup that the system needs to evaluate.

\begin{itemize}
\item {\bf Historical Jobs}: We include all the jobs that are enqueued into the system in the simulation along with their metadata.

\item {\bf Job Metadata:} This includes additional information about each job such as their source (e.g. whether they are user  or automated reports of violation), their violation classifier score. Since this is historical data, jobs also have their actual labels associated with them, which is an important part of the metadata. Finally, we also include the handle time of the job in the metadata, i.e. the amount of time the reviewer spent on reviewing the job.

\item {\bf Job trajectory:} The trajectory of each job is faithfully replicated in the simulator. Jobs get transferred across queues, skipped, reviewed, decisioned, reopened, etc.  As an example, consider a queue that contains jobs that are videos. Videos might require consistent monitoring at frequent intervals to ensure that they contain no harmful content. Such jobs are ``paused'' and put back into the queue and would be ``reopened'' at a later time interval. Similarly, if there are no reviewers that are skilled to work on a particular job (because of, e.g., a language barrier), such a job would be transferred across different queues.

\item {\bf Content-view trajectory:} Since the amount of harm of violating content depends on the number of content views, our simulations include temporal trajectories of content-views for the jobs being simulated. Our systems also use machine learning models to predict content-views, and these prediction trajectories are also included in the simulations.

\begin{figure}
  \centering
  \begin{subfigure}{\columnwidth}
    \centering
    \includegraphics[width=\columnwidth]{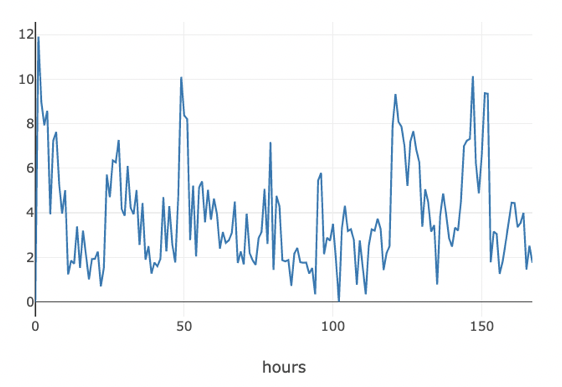}
    \caption{Reviewing capacity as a function of time}
    \label{fig:rep_hours}
  \end{subfigure}

  \vspace{1em}

  \begin{subfigure}{\columnwidth}
    \includegraphics[width=\columnwidth]{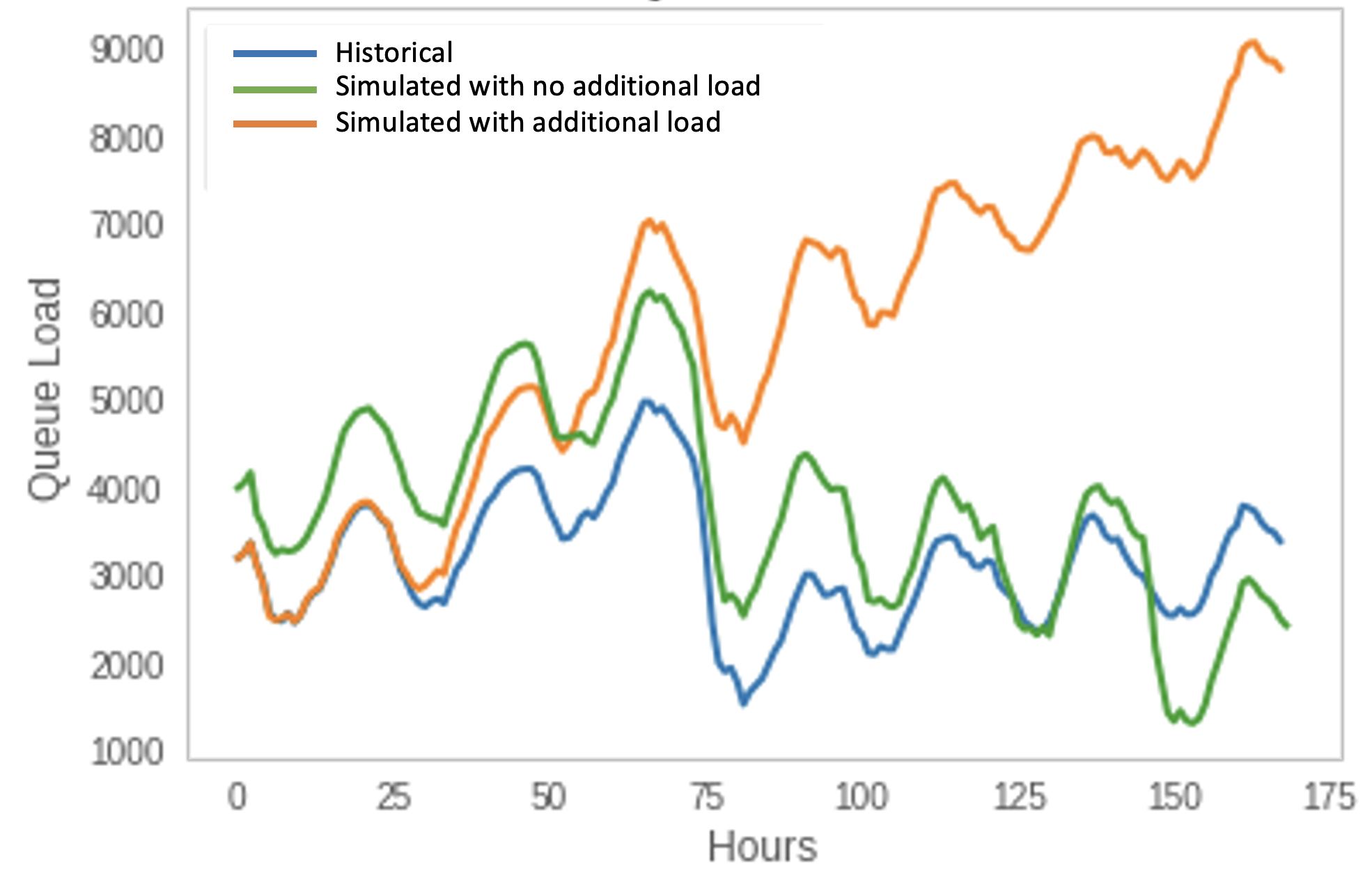}
    \caption{Queue load as a function of time}    
    \label{fig:queueload}
  \end{subfigure}

  \vspace{1em}

  \caption{Example of input historical data (reviewing capacity) and output evaluation metrics (queue load) over time. Notice how queue load tracks  reviewing capacity as expected.}
\end{figure}

\item {\bf Capacity and reviewer schedules}: Reviewers log in to the review system when they are ready and log out when they finish or they are unavailable. This information provides a fine grained view of the reviewing capacity at any point of time. See Fig.~\ref{fig:rep_hours} for an example.

One of the key challenges is that of measuring the productive time of reviewers because; productive hours often do not tally with total work hours. In general a work day comprises of several periods of inactivity.
Times of inactivity can be approximately quantified with proxies such as when no mouse or keyboard activity is detected. Another way to account for this is through the total review time, defined as the total time between a job is rendered on the screen of the reviewer and a job is decisioned minus the inactive time.

\item{\bf Reviewer skills:} Reviewers are typically associated with skills, i.e. attributes that make a reviewer eligible to work on certain subset of jobs (but not others). A reviewer may only work on a job if she is skilled to do so. In order to faithfully simulate our system, we extract the skilling information and include it in the simulation input.

\item {\bf Work Delivery Configurations}: The production logic that is invoked to assign jobs to reviewers involves a number of configured settings. Our simulations extract these configurations and feed them as input to the simulations.

\item {\bf Simulated changes in Demand/Supply}: In situations where we want to simulate changes in demand (i.e. increase/decrease in job volumes enqueued) or supply (more/fewer reviewers available), we re-sample our historical data to achieve the target demand/supply and use the re-sampled data as input to the simulation. The sampling is done in a way that job characteristics (review time distributions and arrival time distributions) and reviewer schedule distributions are roughly maintained the same. These simulations also help us understand how robust our systems are especially when the demand increases (say during event of an election) or when the supply would decrease (reviewers with a particular skill set are not available). 

\end{itemize}

\subsection{Simulation Design}

\nicocom{Section feels too light when this is the core of what we've done. Are there really no design decisions in simpy that one can comment on? is there a single way to implement this in that library that is straightforward to anybody who knows how to use simpy? In addition: figure still says vpvs}
There are several challenges in designing a simulator that can approximate the aforementioned complex interdependencies associated with a real-world queuing system. In particular, a realistic simulator should account for multiple reviewers and queues and job trajectories that involve jobs being re-inserted back into queues or being transferred out. Furthermore, the simulator should be flexible enough to model reviewer schedules and temporal queue loads, whose virality changes as jobs wait in a queue. 

\begin{figure}[tb]
   \hspace*{-5mm}
   \includegraphics[width=.51\textwidth]{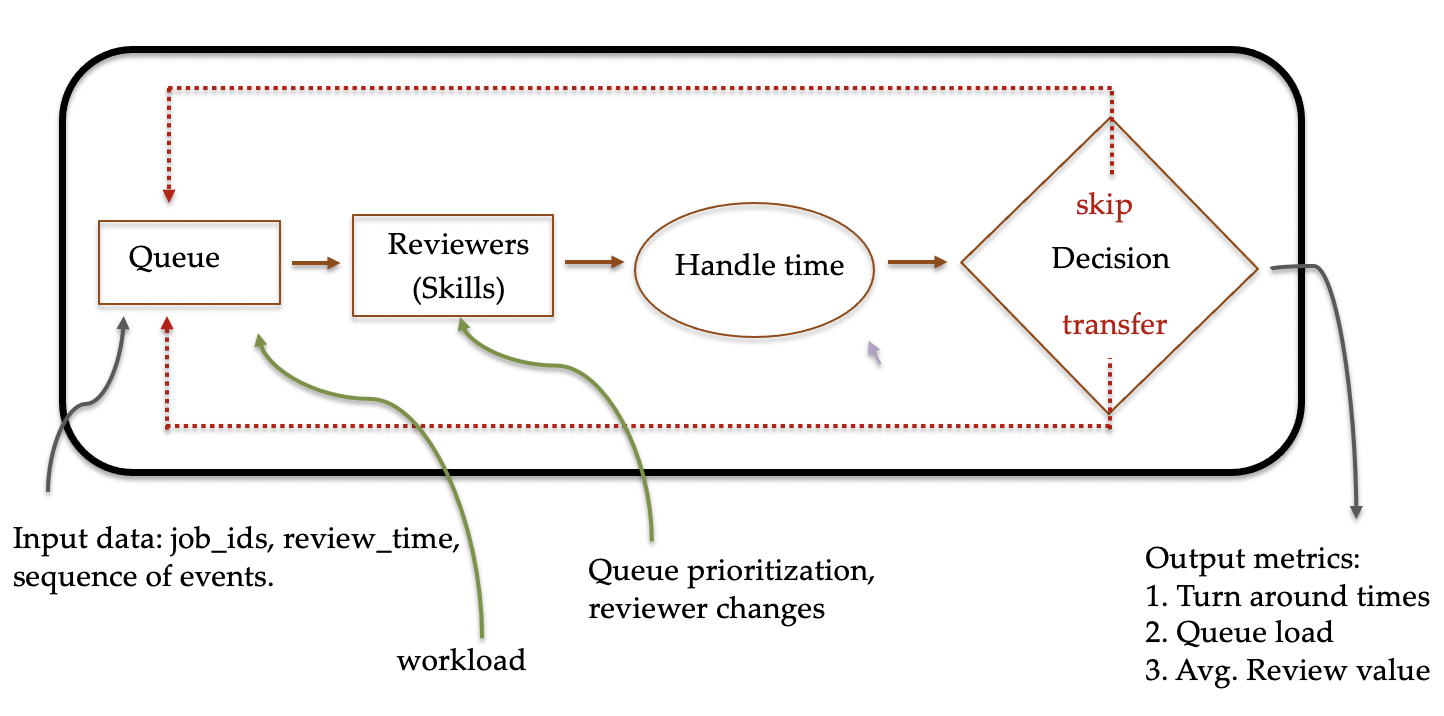}
   \caption{Overview of the simulator}
   \label{fig:simulator}
 \end{figure}

In order to capture the salient features of Facebook's content moderation system, we implemented the discrete-event simulation on top of the Python's {\tt simpy} library \cite{simpy}. This allows one to model processes and objects that interact with one another. More specifically, we define each queue and reviewer as a separate process. In a queue process, we maintain a priority queue where jobs are added as they are created and stored in the order of importance. In the reviewer process, we constantly check for any available jobs to be reviewed in the queues that a specific reviewer is eligible to work and action upon the highest priority job. In cases, where a job needs to be transferred to a different queue or reinserted back into the same queue  Furthermore, we also enable schedules for reviewers by pausing this process when the reviewer's unavailable and restart the process when the reviewer returns back to work. Fig.~\ref{fig:simulator} describes the simulation at a high level. All these events in the simulation are based on a meticulously prepared input data set that attempts to reasonaly approximate anonymized real events. Within the simulation, we maintain logs of the events that can be used compute relevant performance metrics.



\subsection{Evaluation Metrics} 
\label{sec:evaluation_metrics}

In the following discussion, we introduce metrics that are pertinent to making counterfactual business decisions of interest.

\subsubsection{Queue Load}

The {\em load} of a queue at time $t$ refers to the number of open jobs in that queue at that time. See Fig.~\ref{fig:queueload} for an example. The quantity of interest when comparing across scenarios is the change in the peak load\footnote{The peak load is the maximum queue load in the time series.} for each queue, because it is intimately tied to service level agreements and processing times.

\subsubsection{Review Value}

The {\em review value} (RV) metric assigns an importance to any particular piece of content review. The RV of a job can be estimated in a severity-aware manner as the rate of bad experiences prevented relative to the cost and effort to review that content. A simple representation of RV might depend on the predicted content views, the severity, and the review time.
The predicted content views themselves are dynamic, and one must account for the same by forming the prediction at the time when the content is reviewed or enforced.
The higher is the severity of a job that is taken, it is more imperative to take an review such content and take action at the earliest to prevent bad experiences. We can think of this problem as a constrained optimization problem where the goal is to maximize RV for a fixed supply of reviewer time.

\subsubsection{Utilization}

Utilization of a subset of reviewers is the ratio of the total of time reviewers spend handling jobs to total amount of time they are available for work in the system. Utilization can be defined over different job attributes like market or violation type.






\subsubsection{Jobs Closed and Jobs Actioned}

Another important metric at the job level is to track the number of metrics that are closed (worked)
and actioned (jobs that actually involve taking down a content, post). 
While these metrics are positively correlated, it is important that note that the reviewer time is spent on harmful content 
lest it involves too much time being spent on content that is benign.






\section{Simulator Use Cases}
\label{sec:usecases}

In this section we discuss some of the operational challenges involved in  running a large scale human reviewing system and how QUEST simulations allows us to tackle them. We do so by focusing on four specific use cases.

\subsection{Capacity Planning}
One of the most important recurrent decisions is to dimension the system properly and configure the assignment of reviewers to queues. We use QUEST to 
forecast the operating variables and make sure that they satisfy the needs and to help optimize the assignment so there is sufficient capacity to review the jobs that enter queues.

\begin{figure}[th]
  \noindent
  \begin{subfigure}{\columnwidth}
    \includegraphics[width=\columnwidth]{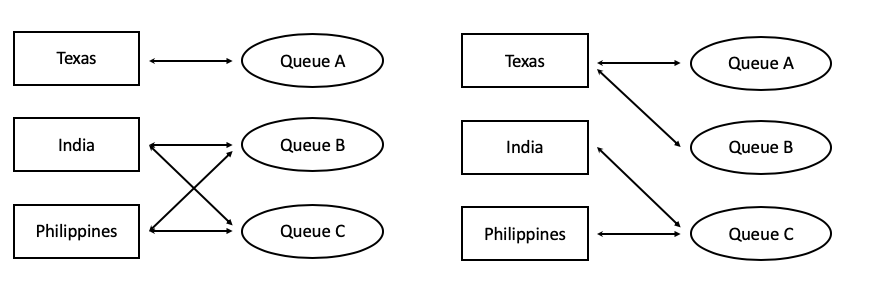}
    \caption{Two different assignments of reviewers to queues}
    \label{fig:capscenario}
  \end{subfigure}

  \vspace{1em}

  \begin{subfigure}{\columnwidth}
    \small
    \begin{tabular}{|c|c|c|c|c|}\hline
        &Reviewer & Job   &           & Avg. TAT\\
        & hours   & hours & Utilization & (sec) \\
        \hline
        Status quo  &  1660 & 1601 & 96.4\% & 189 \\\hline
        Move queue $B$  & &&&\\
        to Texas & 1660 & 1653 & 99.6\% & 1343  \\\hline
    \end{tabular}
    \caption{
      Effects on TAT of when the utilization in Texas is increased. 
    }
    \label{fig:move_to_texas}
  \end{subfigure}

  \vspace{1em}
  \caption{Capacity planning use case: Establishing the effects of a potential re-assignment of reviewers to queues. Reviewer hours and job hours are based on historical data.}
\end{figure}

As an example, consider the following small instance with queues $A$, $B$ and $C$ and reviewers working from three locations: Texas, India and Philippines (see Fig.~\ref{fig:capscenario}). In the initial scenario reviewers from Texas are assigned to jobs from queue $A$ while reviewers from India and Philippines are assigned to jobs from queues $B$ and $C$. A new configuration is proposed under which queue $B$ would be reassigned to Texas, e.g., because of time zone considerations. 
Before implementing the proposal in production, one must be convinced that it is going to improve the status-quo. 
\begin{itemize}
\item Are there tangible benefits and improvements in accuracy?
\item How would the TAT be impacted?
\item Does the Texas center have enough capacity to handle the additional load? If not, how many additional reviewers should we hire?
\end{itemize}
A QUEST simulation can be used to readily answer these questions. For example, in Fig.~\ref{fig:move_to_texas} we see that the proposed change has a significant adverse effect on the avg. TAT of jobs in the queue. 

Besides stationary results, it is also very important to measure the dynamic behavior of queues when launching a new configuration. For example, consider that we need to add $200$ hours of content moderation to a queue. To address that, we can create the equivalent of $200$ hours worth of synthetic jobs within the simulation.
As one set of inputs that we mentioned earlier, QUEST has the ability to run an impact analysis for ad-hoc scenarios that require one off injections of volume to specific queues. 

\subsection{Threshold Generation for Validation of Proactive Reviews}
\label{subsec:automation}

Classifier training and reviewer accuracy measurement are a vital component of content moderation. Machine learning classifiers are used extensively to proactively detect content. According to the Facebook Community Standards Enforcement Report, due to investments in AI, Facebook has been able to remove more hate speech and find more of it proactively before users report it and
about 95\% of hate speech content was proactively identified \cite{FacebookCSERnov20}. 

One of the crucial steps in operating the proactive system is to review borderline content by human reps, and to validate the accuracy of the classifiers. One of the ways one might address this is by sending a sample of the content proactively identified by classifiers 
to the human labeling queues described earlier.

We used QUEST to simulate a dual review system based on both machine learning and human labeling, and to find an optimal operating points in a case study.
We used criteria to trigger the validation that depend on the content type, the content views since posted, and on whether a classifier-generated violation score is under a corresponding threshold. A job-reaper was set up in the simulation to periodically check content and automatically enqueue those satisfying these criteria. 

Since the effective use of human review capacity depends on the choice of thresholds for all markets and queues, we rely on simulations for different parameters. This allows us to achieve the right trade-offs in the output metrics. 
To answer these questions,
we collect historical job-level data to create a view of a market within a specified time-window. Within this view, we collect all the relevant information about the job, such as its TAT, its latent content view, its classifier violation score, if automation was used, whether it was violating, and its severity level. We can then replay the dataset via simulations with varying criteria.

As an example, suppose we want to evaluate the effect of changing the threshold for the violation score from $x$ to $y$. The effect of this change could be evaluated by understanding the resulting additional enqueued jobs and the change in relevant output metrics.

\begin{figure}
  \centering
    \centering
    \includegraphics[height = 5cm, width=0.9\columnwidth]{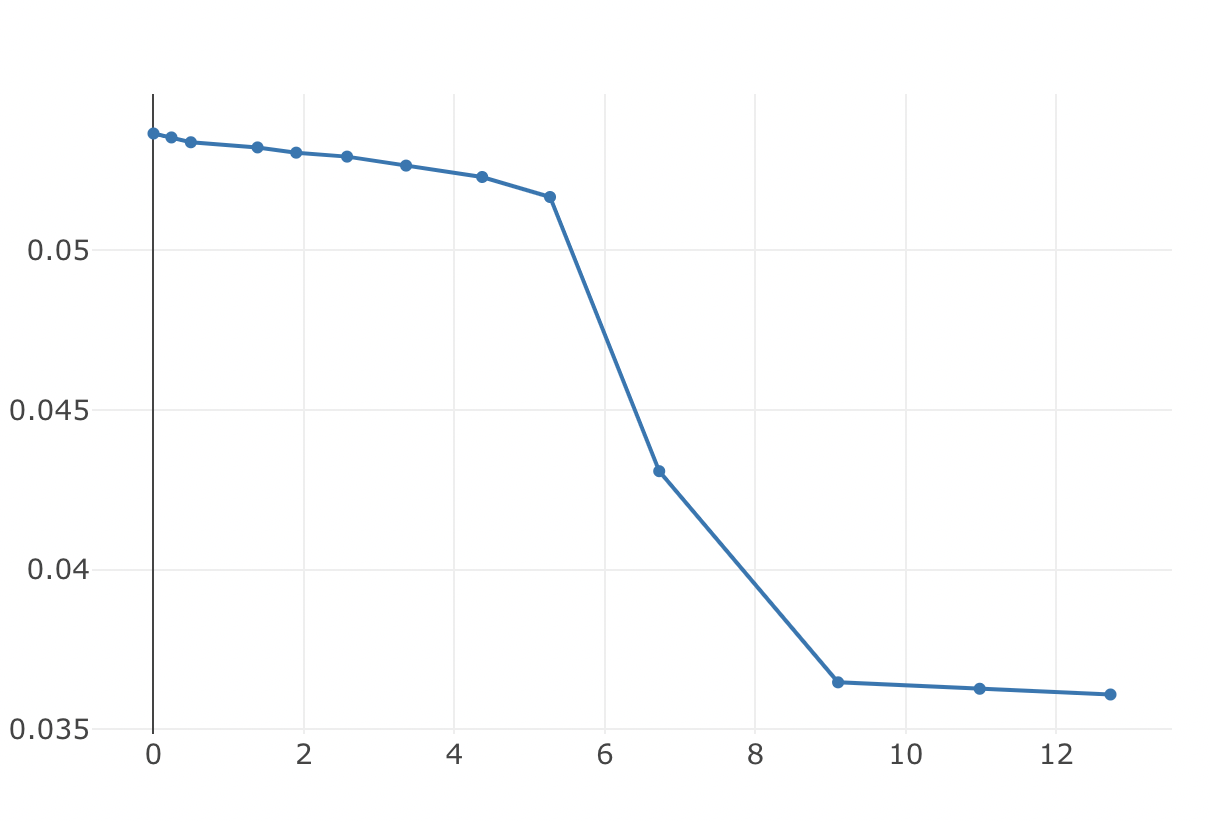}
    \caption{Simulated trade-off curves between fraction of the data used for validation vs.\ misspecification of violating content.}
    \label{fig:vhs-miss-rate}

  \label{fig:segment}
\end{figure}

In Fig.~\ref{fig:vhs-miss-rate}, we quantify the validation for high severity content as a function of percentage of the total jobs that were used for validation in our case study. As the validation sample gets larger, we see a decrease in misspecification rate. 

\subsection{Intra-Queue Job Prioritization}
\label{subsec:job_queue_priotization}

Queues are classified into segments that capture the job type such as severity, content type, etc. 
Whenever a reviewer becomes available, the job allocation system first chooses a segment and then considers the available queues for that reviewer within that segment, and picks a job from one of them. 
Indeed, to assign jobs, the system ranks them among the queue, ranks the queues within segments and ranks the segments among themselves. 

We now describe the intra-queue prioritization (job ranking within a queue)
and discuss the inter-queue prioritization (ranking among queues and segments) in Sect.~\ref{subsec:inter_queue_priotization}.

Queue prioritization is done through a value model that we refer to as {\em prioritization formula}. Some simple examples include FIFO, Smith's rule (bang-per-buck), linear combinations of relevant features, to more complex combinations of features and arbitrary ML models. The prioritization is a crucial lever that ensures that the review capacity is leveraged to the best possible extent by reviewing jobs in the correct order. The system can dynamically reorder the jobs as more information becomes available.
The main parameters that need to be configured are the number of jobs in the queues and the reorder frequency.
Nevertheless, there are several factors that make the design of a queue prioritization formula challenging. We describe some common situations for which QUEST can be used to help improve the system.
\begin{enumerate}
\item {\bf Evaluating cost vs.\ benefit}: Starting with a current prioritization formula, one may wish to estimate the possible gains of adopting an optimized one. The goal would be to decide if the gains outweigh the extra development costs and ensuing engineering complexity. 
\item {\bf Balancing trade-offs}: It is common to have several metrics of interest. Choosing the right prioritization formula involves understanding trade-offs between them, and choosing an operating point in the Pareto frontier.
\item {\bf Parameter sweeps}: The queue prioritization framework may involve parameters that need to be tuned. These parameters could be part of the queue prioritization formula, or could be define other aspect of the operations of the system such as the maximum queue size, the reorder frequency, automation thresholds as described earlier, etc. QUEST can be used to 
sweep the parameter space and find a good operating point looking at the predicted output metrics. 
\item {\bf Impact of classifiers}: It is possible for the prioritization formula to depend on signals computed by ML classifiers. QUEST can also be used to pick or tune those ML models.
\item {\bf Operational impact}: In some cases, we may want to answer questions that involve counterfactuals with the data. As an example, QUEST can compute how much reduction of bad content one could get if we increased review capacity by a given percentage in a market. Since queue prioritization critically affects the impact, capturing its effect in the estimation is important. 
\end{enumerate}

The next sections discuss two concrete use cases that introduce additional details about optimizing the intra-queue prioritization process.

\subsubsection{Optimizing TAT}

\begin{figure}
  \begin{subfigure}{\columnwidth}
    \centering
    \includegraphics[width=0.9\columnwidth]{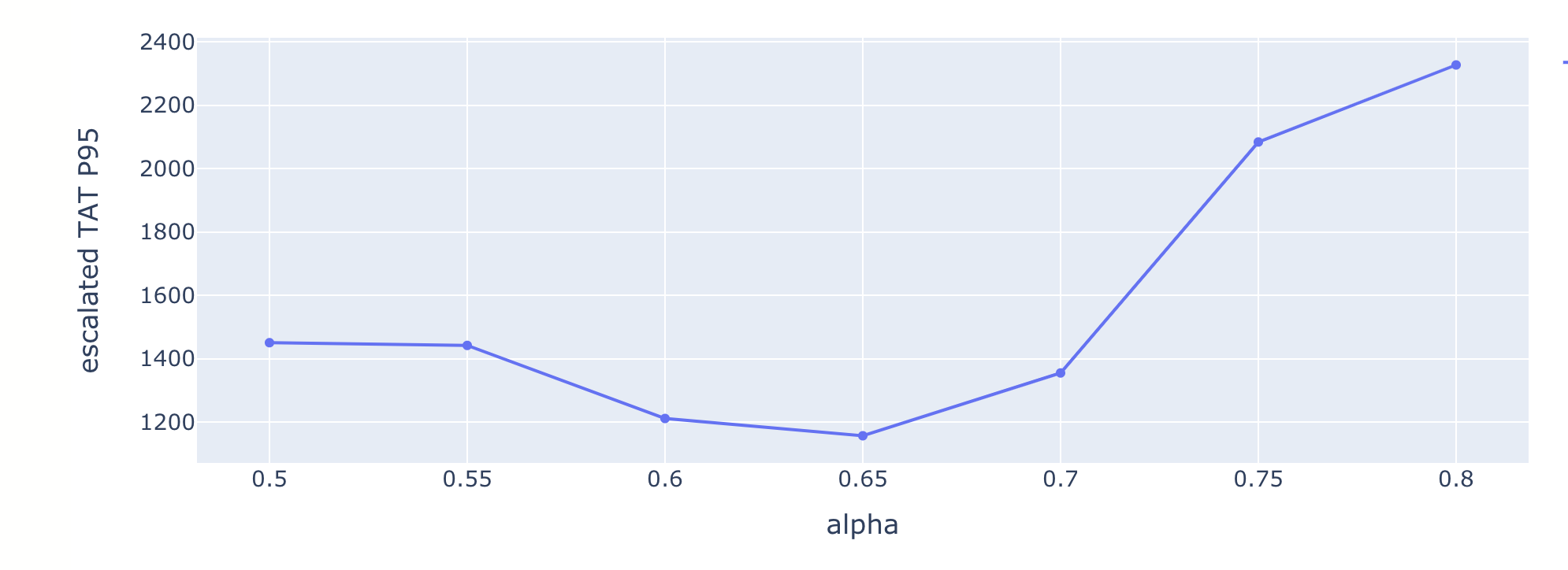}
    \caption{TAT for escalated jobs as a function of $\alpha$}
    \label{fig:queuepri3}    
  \end{subfigure}

  \vspace{1em}

  \begin{subfigure}{\columnwidth}
    \centering    
    \includegraphics[width=0.9\columnwidth]{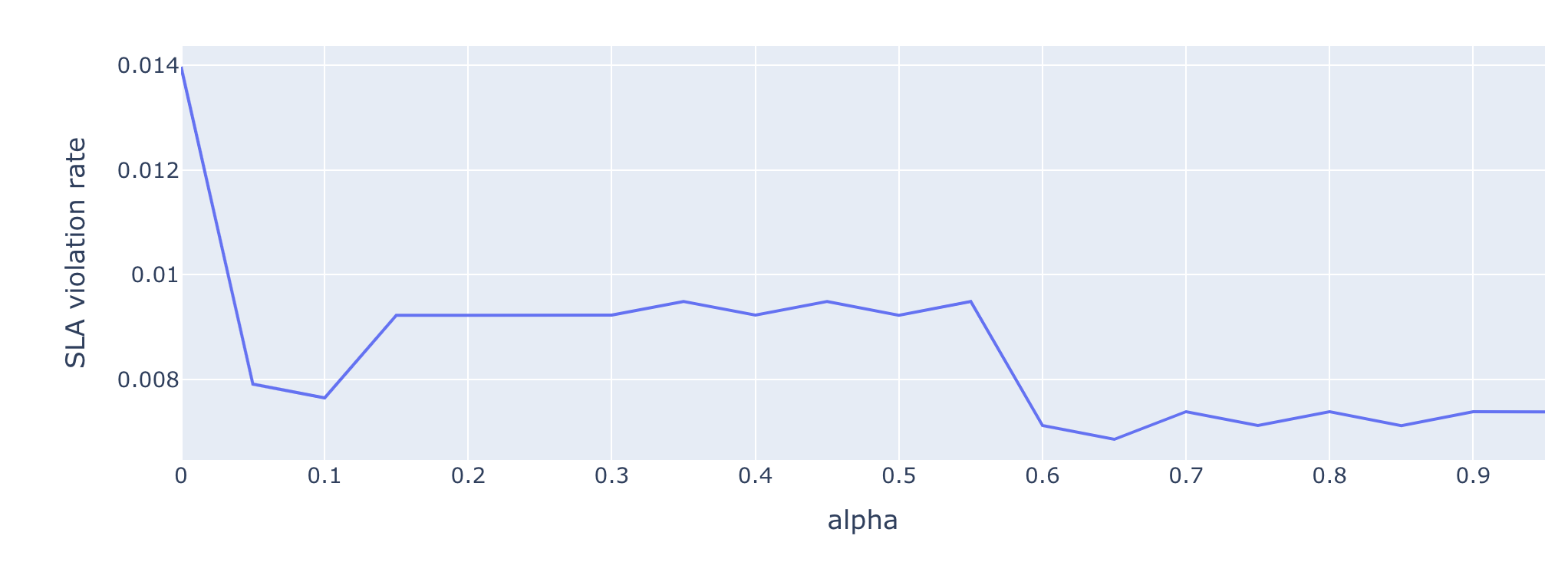}
    \caption{SLA violation rates as a function of $\alpha$.}
    \label{fig:queuepri4}      
  \end{subfigure}

  \vspace{1em}

  \caption{Intra-queue prioritization use case: Investigating the effect of parameter $\alpha$ to competing metrics. As expected, as we increase $\alpha$, TAT for escalated jobs increases, while SLA violation rate decreases.}

\end{figure}

Certain violation reports require very small TATs. To ensure this, we considered a  queue prioritization formula for the corresponding queues that was better tuned to minimize the expected TAT of escalated jobs to make sure that flagged content was promptly reviewed. Besides the expected TAT, all jobs in the queue have to be reviewed within a time window to guarantee an SLA. The proposed prioritization formula was a function of $p_{\text{escalate}}$, estimated by a classifier that predicts the probability of an escalation event. 
The proposed queue prioritization formula was
$$
p(t) = p_{\text{escalate}} + \alpha \cdot \text{commit}(t),
$$
where $\text{commit}(t)$ is a function that increases with the time a job has been enqueued.
Figs.~\ref{fig:queuepri3} and~\ref{fig:queuepri4} demonstrate how the simulator may be used to balance the trade-off between optimization of two competing metrics (TAT for escalated jobs vs. SLA) and tune the queue prioritization rule via the parameter $\alpha$.


\subsubsection{Virality Monitoring}

Virality is a phenomenon of key importance.
Content that rapidly accumulates content views poses a higher risk if violating because it can be exposed to a broad audience before it can be taken down. Consequently, it is important to monitor virality. To address this, the system could review most of the content on the platform achieving a distribution above a high percentile. One of the key metrics of interest is coverage among that subset of the content (i.e., fraction reviewed among posts with high distribution). This can be defined for all content or for specific content types. As an example, we consider three content types of particular interest: civic-, health- and COVID-related content. We have access to classifiers that predict the probability of content being of a particular type, these are captured by the quantities $p_1, p_2, p_3$. For this, we consider the queue prioritization formula:
$$
s = \text{content-view}_\text{pred} \cdot \max(1,w_1\cdot p_1, w_2 \cdot p_2, w_3 \cdot p_3).
$$

\begin{figure}
  \centering

  \begin{subfigure}{\columnwidth}
    \centering
    \includegraphics[width=0.9\columnwidth]{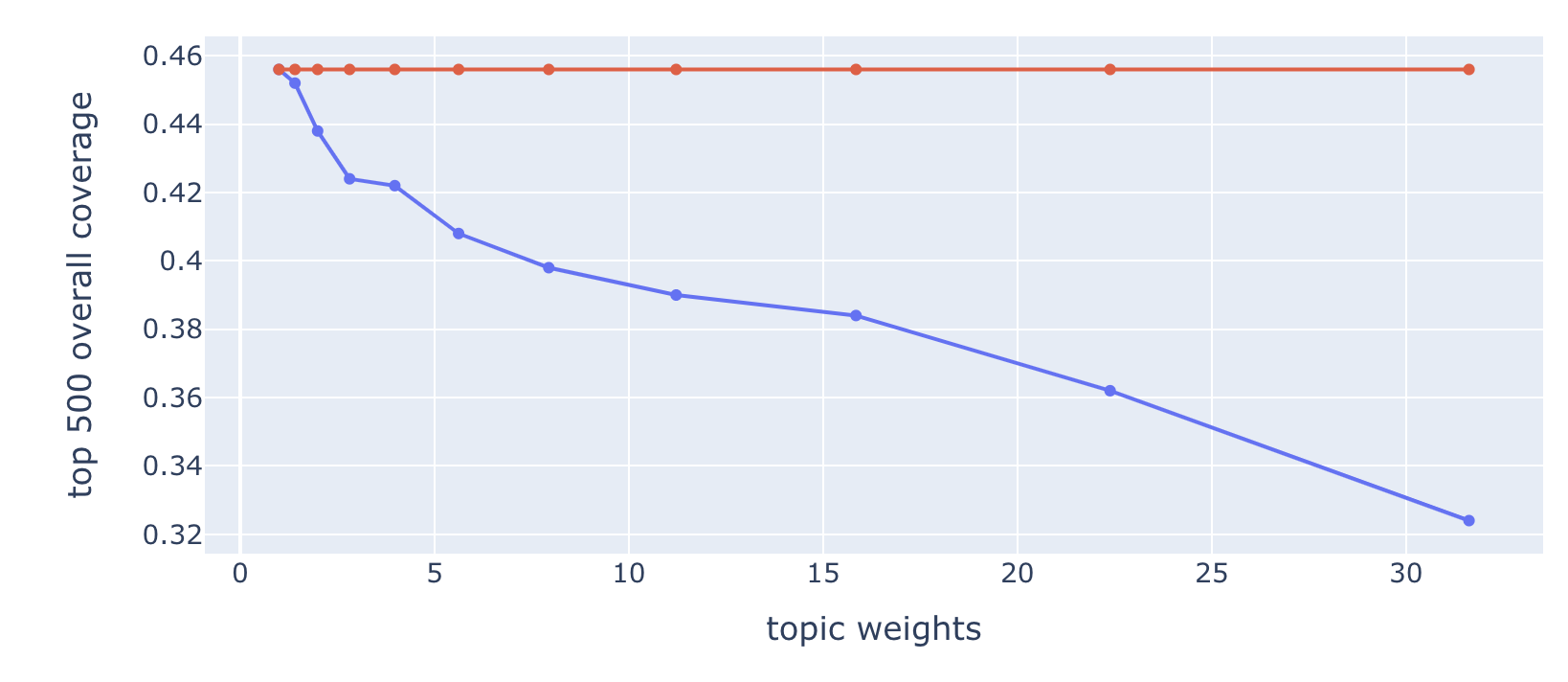}
    \caption{Overall coverage as a function of weight.}
    \label{fig:queuepri1}
  \end{subfigure}

  \vspace{1em}

  \begin{subfigure}{\columnwidth}
    \centering
    \includegraphics[width=0.9\columnwidth]{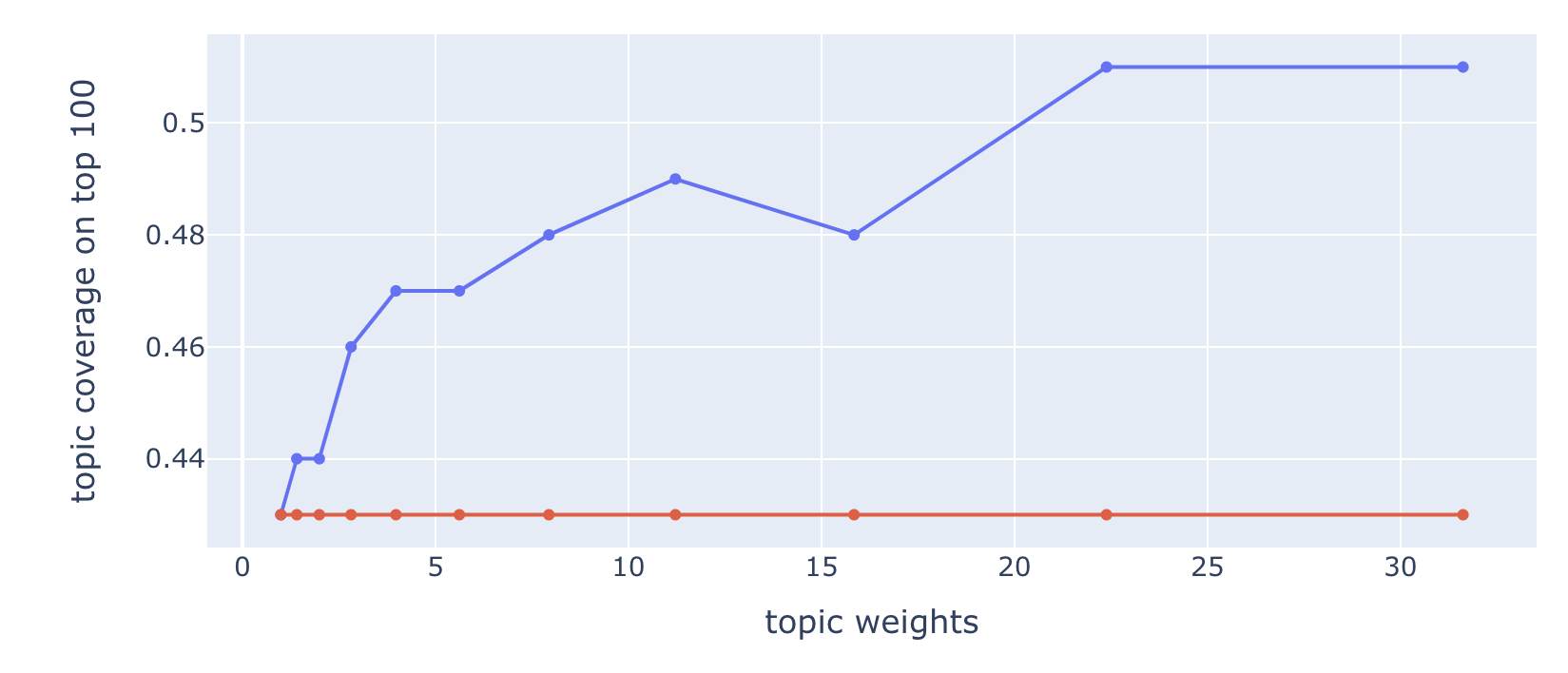}
    \caption{Topic coverage as a function of weight.}
    \label{fig:queuepri2}  
  \end{subfigure}

  \vspace{1em}

  \caption{Intra-queue prioritization use case: As we increase the total weight of the topics we want to prioritize, their coverage goes up at the expense of the overall content coverage. The constant baseline corresponds to ranking by predicted content views alone.}
\end{figure}

The priority of a job is defined to be a function of the predicted content views accrued (based on content-view prediction models), as well as classifier signals for civic, health and COVID-related content. The parameters $w_1$, $w_2$, $w_3$ are used to boost the relative importance of one type of signal against others. When $w_0$ is boosted up we optimize for overall coverage, when $w_1$ is boosted, we optimize for coverage of civic content, and so on. We ran a suite of simulations where we held $w_0$ = 1 fixed (since only the relative scaling matters), and set $w_1=w_2=w_3=W$. We swept over the parameter $W$ to understand the impact on coverage. The simulation plots in Figs.~\ref{fig:queuepri1} and~\ref{fig:queuepri2} confirm our intuition that as we increase the total weight of the topics, their coverage goes up at the expense of overall content coverage. Precise trade-offs curves like these allow decision makers to answer questions like who much can we afford to boost these topics without decreasing overall coverage more that a given amount; e.g., the plots suggest that by setting the weight to a relatively low value (say $W=4$) we can obtain at least half of the possible gains in topic coverage only at a modest decrease in overall content coverage.

\subsection{Inter-Queue Prioritization}
\label{subsec:inter_queue_priotization}

There are multiple ways to rank queues and segments. Similarly to the value model described earlier, one possibility is that the system prioritizes the queues and segments according to their importance, which is encoded in a formula. Jobs are selected from a particular segment until the capacity allocated for that segment is exhausted, as it is done for queues. This could easily lead to a queue being starved if the reviewers that can work on the queue are also able to work on another queue with higher priority.




Another option instead of the prioritization model is to use {\em percentage based allocations} to sample the queues. We illustrate this describing the case of two queues. Consider queues $A$ and $B$, and a percentage allocation $x:y$. If both queues have jobs, we select a queue with a probability proportional to $x$ or $y$, and pick the most important job in that queue. The percentages can be prefixed, adjusted based on daily targets and total hours of  work in the segment and queue, or tuned dynamically based on the available information.

Note that as with the intra-queue prioritization, the system could use different strategies for different segments or queues.

\subsubsection*{Simulation Results under different allocation strategies}

\begin{figure}
  \centering
  \begin{subfigure}{\columnwidth}
    \centering
    \includegraphics[height=6cm, width=0.9\columnwidth]{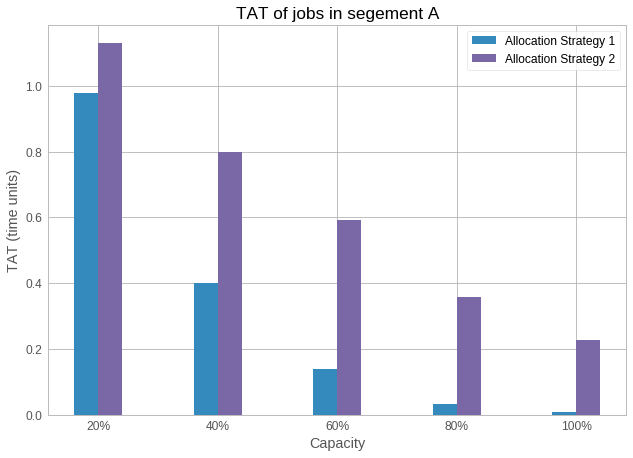}
    \caption{TAT of jobs in segment A as the capacity of reviewers is varied}
    \label{fig:tat_allocation}
  \end{subfigure}

  \vspace{1em}

  \begin{subfigure}{\columnwidth}
    \centering
    \includegraphics[height=6cm, width=0.9\columnwidth]{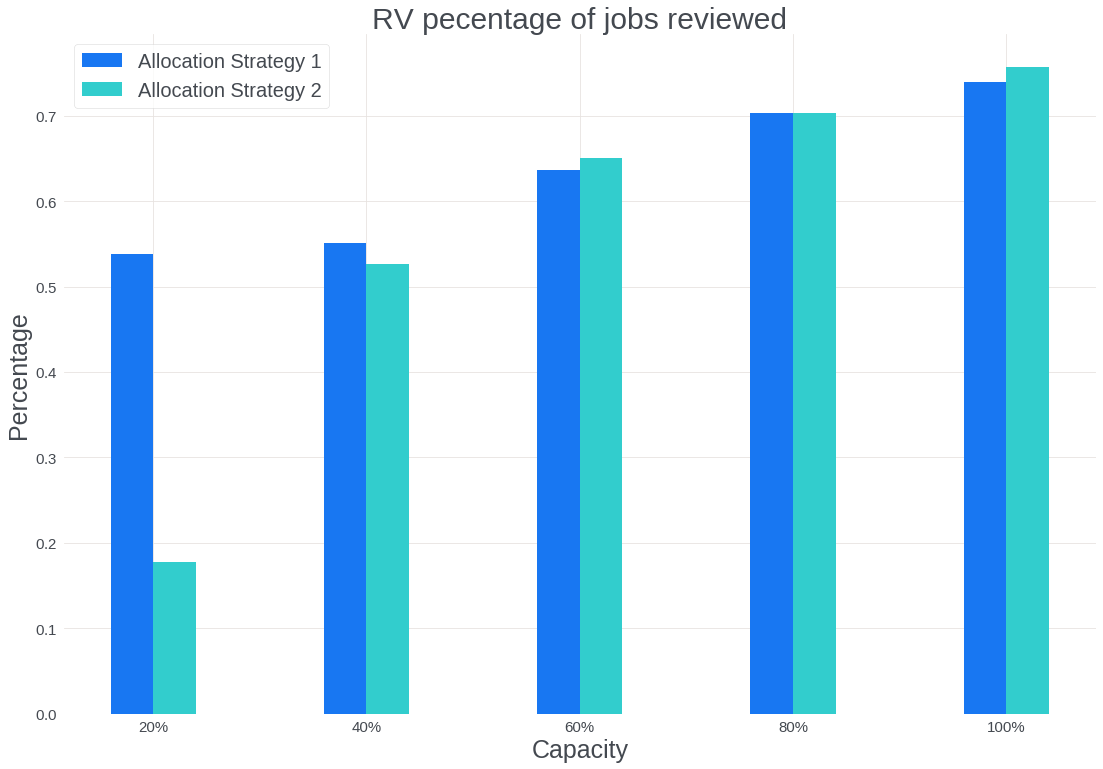}
    \caption{Percentage of RV for different capacity levels}
    \label{fig:irv_percentage}
    \end{subfigure}

    \vspace{1em}

    \caption{Inter-queue prioritization use case: Investigating the effects of two allocation strategies for ranking segments and queues.}
\end{figure}

We used QUEST to help with the choice between the two different allocation strategies described above. We consider three segments $\{A, B, C\}$ over 51 queues in a market. Most of the jobs with high RV are present in segments $A$ and $C$. The content views of a job change as a function of time. 
We also have classifiers that predict the RV in advance that can be used to decide which jobs must be enqueued. The accuracy of RV depends on the accuracy and calibration of content view predictions and review time predictions. The allocations strategies are:
\begin{itemize}
\item \textbf{Allocation Strategy 1}: We stack rank segments $A$, $B$ and $C$. The queues in each segment are further stack ranked amongst themselves within inside segment. 
\item \textbf{Allocation Strategy 2}:
We combine segments $A$ and $C$ and run a percentage base allocation of $60\%:40\%$ between the combination of segments $A$ and $C$ vs.\ segment B. In this model when a reviewer requests a job, they have a $40\%$ chance of getting a job from segment $B$, and $60\%$ chance of getting a job from segments $A$ or $C$. Within the combined segments $A$ and $C$, we first pick jobs from queues in segment $A$ and then~$C$.
\end{itemize}

 The two metrics that we used to compare these strategies on were the total review value of jobs that are reviewed and the TAT of jobs in segment A. Figs.~\ref{fig:tat_allocation} and~\ref{fig:irv_percentage} show the outputs of the simulations. We also look at these metrics under different supply capacity levels. We vary the review capacity that is available and look at which strategy leads to higher review value.  As expected as the review capacity increases, the TAT of jobs in segment $A$ decreases for both allocation strategies. However the decrease is much more prominent with stack ranking as the jobs from segment $A$ are given more priority (jobs from that segment are worked first). At lower capacity levels, strategy 1 leads to higher review value as predominantly most of the review capacity is directed towards working jobs from segment A, however as review capacity is increased and jobs from segment C are worked, strategy 2 leads to a higher review value. Such simulations help us make trade-off decisions among various allocation strategies.

\section{Conclusion \& Future Work}
\label{sec:conclusion}

We have introduced the QUEST discrete simulation model which can be leveraged to optimize content moderation systems. In particular, we have described how simulations were helpful in evaluating and adopting theoretical ideas to address a number of real-life operational challenges including capacity planning and queue prioritization. In the future, we will build on this work to further leverage existing theoretical frameworks and model additional challenges with simulations to continually improve these systems. 

One improvement that will be interesting to look at is to identify specific reviewers who have previously taken good decisions for individual pieces of content of the same type. It is typical for different reviewers to have different skill sets, ranging from familiarity with a language to expertise in specific content types. Therefore, it is important to optimally match a potentially harmful content with the best reviewer available. However, there is a trade-off between matching content with best reviewers and having a quick TAT. While it is possible to develop principled heuristics based on online optimization or back pressure methodologies \cite{backpressure}, it is not easy to quantify these trade-offs owing to the dynamic nature of content moderation. We plan to use the simulator to analyze these methods and quantify the gains from such optimization. 

Furthermore, the simulator can be useful in quantifying the trade-offs in improving the accuracy of ambiguous or borderline content, where we need to rely on additional reviews to reduce potential mistakes. 

In summary, QUEST is a powerful tool that can help us in iterating through many principled ideas faster and quantify the potential improvements in implementing these ideas in production.

\section*{Acknowledgements}

A lot of this work would not have been possible without the help of our collaborators and colleagues at Facebook. We would specifically like to acknowledge (in no particular order)  
Ajay Menon, Dima Karamshuk, Thomas Leeper, Evgeniy Riabenko,  Milan Vojnovic, Alex Nikulkov, Eric Wang, Vaggos  Chatziafratis, Birce Tezel, Yue Shi, Udi Weinsberg, Aisling McCauley, Sayash Kapoor, Sagnik Ghosh, and Omar Abdul Baki. 

\bibliographystyle{ACM-Reference-Format}
\bibliography{references}

\end{document}